\documentclass[runningheads]{llncs}
\usepackage[T1]{fontenc}
\usepackage{graphicx}
\usepackage{booktabs}
\usepackage[misc]{ifsym}
\newcommand{\corr}{(\Letter)}
\usepackage{amsmath}
\usepackage{amsfonts}
\usepackage{algorithm}
\usepackage{algpseudocode}
\usepackage{multirow}
\usepackage{xcolor}
\begin{document}

\title{\textit{DispaRisk}: Assessing Fairness Through Usable Information}

\titlerunning{\textit{DispaRisk}: Assessing Fairness Through Usable Information}
% If the full title of your paper is short enough to also fit in the running head, you can omit the abbreviated paper title here. You can check as follows: if you comment out the \titlerunning line, something will appear in the header of all odd-numbered pages of your PDF from page 3 onward. This something is either the full title (in which case all is well), or the error message "Title Suppressed Due to Excessive Length". If this error message appears, you're going to want to provide an abbreviated title within the \titlerunning command, because if you won't do it, Springer will do it for you.

%N.B.: Author information (both in the \author{} and \authorrunning{} command) should only be present in the Camera-Ready Version of your paper. The version that you initially submit for review, ought to be double-blind. So, when initially submitting your paper, use:
% \author{Author information scrubbed for double-blind reviewing}
\author{Jonathan Vasquez\inst{1,2}\corr \and
Carlotta Domeniconi\inst{1}  \and
Huzefa Rangwala\inst{1}}
% You may leave out the orcidID information, if you want to.
% Use \corr to indicate the corresponding author. Note the spacing around the \corr command. Only one author can be the corresponding author.

%N.B.: comment out the \authorrunning{} command for the double-blind version of your paper submitted for review. Later, if your paper is accepted, use the command for the Camera-Ready Version.
\authorrunning{J. Vasquez et al.}
% If there are two authors, write 'A.L. Benjamin and C.C. Broadus Jr.'
% If there are more than two authors, '[...] et al.' is used.

\institute{George Mason University, Fairfax, USA 
\email{\{jvasqu6,cdomeniconi,rangwala\}@gmu.edu}
\and
Universidad de Valparaiso, Chile\\
\email{jonathan.vasquez@uv.cl}
}
\maketitle              % typeset the header of the contribution

\begin{abstract}
Machine Learning algorithms (ML) impact virtually every aspect of human lives and have found use across diverse sectors including healthcare, finance, and education. Often, ML algorithms have been found to exacerbate societal biases present in datasets leading to adversarial impacts on subsets/groups of individuals and in many cases on minority groups. To effectively mitigate these untoward effects, it is crucial that disparities/biases are identified early in a ML pipeline. This proactive approach facilitates timely interventions to prevent bias amplification and reduce complexity at later stages of model development. In this paper, we leverage recent advancements in usable information theory to introduce DispaRisk, a novel framework designed to proactively assess the potential risks of disparities in datasets during the initial stages of the ML pipeline. We evaluate DispaRisk's effectiveness by benchmarking it against commonly used datasets in fairness research. Our findings demonstrate DispaRisk's capabilities to identify datasets with a high risk of discrimination, detect model families prone to biases within an ML pipeline, and enhance the explainability of these bias risks. This work contributes to the development of fairer ML systems by providing a robust tool for early bias detection and mitigation.

\end{abstract}

\section{Introduction}
Extensive research on fairness in machine learning (ML) highlights how biased datasets can amplify historical and societal biases \cite{pessach2023algorithmic,kizilcec2022algorithmic,wen2021algorithms,fu2021crowds}, harming minorities and disadvantaged groups in areas like criminal justice \cite{ProPublica2016}, healthcare \cite{chen2023algorithmic}, and education \cite{vasquez2022faired}.. This underscores the need to detect biases throughout the ML pipeline—especially in its early stages \cite{feldman2015certifying,hardt2016equality}.  To address this, \textit{data-} and \textit{model-focused} metrics help identify potential discrimination risks, though they have limitations:

\textit{Data-focused} metrics are computed directly from the dataset and include Class Imbalance (CL) \cite{hardt2021amazon}, Difference in Positive proportions in observed Labels (DPL) \cite{hardt2021amazon}, and Mutual Information (MI) between the sensitive attribute and the rest of the features \cite{gupta2021controllable}. While useful, these approaches do not account for model selection or preferences, making it difficult to determine which models are more likely to produce disparate outcomes. To address this gap, \textit{model-focused} metrics analyze trained models directly. These methods, discussed by \cite{hardt2021amazon,vasquez2022faired}, detect existing discrimination (e.g., Demographic Disparity (DEMP) and Equalized Opportunity (EQODD)) or provide explanations for disparate predictions (e.g., KernelSHAP \cite{lundberg2017shap,shapley1953values}). However, these evaluations occur late in the pipeline and are tied to specific models, limiting their generalizability. Moreover, they do not fully capture how the interaction between data characteristics and model capabilities affects fairness.

While fairness metrics help assess bias at different stages, they do not address how models interact with data properties in practice. Even when datasets appear balanced, models may process information differently across groups, leading to hidden disparities. For instance, in an ML pipeline classifying rural and urban loan applicants as approved or denied, a dataset may predict approvals equally across groups. However, simple models might effectively leverage credit scores for urban applicants while struggling with interaction-based features critical for rural ones. This disparity in information usability can lead to uneven model performance and potential discrimination, even with seemingly fair datasets. Moreover, increasing model complexity does not necessarily resolve these issues, as it depends on whether the model can effectively utilize nuanced information for different groups. Hence, key questions emerge: Can differences in usable information across groups be quantified to trace disparities? How does model choice influence these differences and outcomes? Addressing these requires a deeper assessment of ML pipelines, beyond dataset balance and final model evaluation. Specifically, an approach that enables early detection of disparity risks while accounting for model-specific characteristics is needed.

To operationalize these insights, we introduce DispaRisk, a framework designed to detect disparity risks early in the ML pipeline while considering the characteristics of the predictive models being used. Building on the \textit{usable information} notions studied by Xu et al. \cite{xu2020theory}, DispaRisk enables proactive fairness assessments by guiding the estimation of usable information-based metrics. Specifically, given a set of potential model choices, DispaRisk facilitates assessment analyses that: (1) can be conducted in the early stages of ML pipelines, (2) account for the predictive families selected, (3) correlate with data- and model-focused fairness metrics, and (4) explain why different model families generate disparate outcomes. This approach serves as an effective predictor of discrimination risks that may emerge later in the pipeline. 

The key contributions in this study are threefold:
\begin{enumerate}
    \item Introduces DispaRisk, a framework leveraging advancements in \textit{usable information} theory for early-stage disparity detection while accounting for predictive model families in ML pipelines.
    \item Bridges the gap between \textit{data}- and \textit{model}-focused bias assessment approaches.
    \item Demonstrates practical applications through experiments across diverse datasets, showcasing its ability to identify high-risk datasets, detect bias-prone model families, and improve bias explainability.
\end{enumerate}

% The subsequent sections are structured as follows: First, the problem setting is formalized. Next, the rationale of assessment through \textit{usable information} is developed. Then, the DispaRisk framework is introduced, followed by experimental results that validate its effectiveness.\footnote{The code for the experiments can be found here: \url{https://bit.ly/ecml2025}} Finally, we conclude with a discussion of the implications of this work and potential avenues for future research.

\section{Basics and Preliminaries}
\label{sec:basics}
\subsection{ML Pipeline Basics}
\label{sec:ml_basics}
Let $X$, $S$, and $Y$ be random variables in the space $\mathcal{X} \times \mathcal{S} \times \mathcal{Y}$, representing the input features, sensitive attributes, and target variable, respectively. An ML pipeline is given access to a dataset $\mathcal{D}_n = {\{x_i, s_i, y_i\}_{i=1}^n \in \mathcal{X} \times \mathcal{S} \times \mathcal{Y}}$ of $n$ instances to learn a mapping function $h: \mathcal{X} \mapsto \mathcal{Y}$ by employing a finite set of possible models $\mathcal{V}$. We assume that there is access to sufficient information about the ML pipeline to identify the set of possible models.

\subsection{Fairness Notion}
\label{sec:fairness_notion}
We examine fairness through independence and separation notions \cite{feldman2015certifying,hardt2016equality,pessach2023algorithmic,vasquez2022faired}. Independence requires the learned mapping function's outcomes to be independent of the sensitive attribute ($h(X)\perp S$), while separation also requires independence, but conditioned to the ground truth ($h(X)\perp S | Y$). Our analysis focuses on positive class of binary classifications, which typically signify favorable decisions with significant social implications. For example, in contexts such as university admissions or loan approvals, positive outcomes (e.g., being admitted or approved) directly influence individuals' opportunities. To evaluate these disparity kinds, we employ the metrics DEMP and EQOPP explained as follows:

\begin{definition}[Demographic Disparity (DEMP)]
Difference in the \textbf{positive rate} of class $k \in Y$ between the advantage ($s$) and disadvantage ($s'$) groups.
\begin{align*}
    \begin{aligned}
    \Delta_{DEMP} (h,S,Y_k)  = P(h(X) = 1 | S=s) - P(h(X)=1| S=s')
    \end{aligned}
\end{align*}
\end{definition}

\begin{definition}[Equalized Opportunity (OPP)]
Difference in the \textbf{true positive rate} of  class $k \in Y$ between advantage ($s$) and disadvantage ($s'$) groups:
\begin{equation*}
    \begin{aligned}
    \Delta_{OPP}(h,S,Y_k)=  P(h(X) = 1 | S=s, Y_k=1) - P(h(X)=1 | S=s', Y_k=1) 
    \end{aligned}
\end{equation*}
\end{definition}

\subsection{Usable information framework}
\label{sec:fairness_evaluation_usable_information}
The \textit{usable information} framework \cite{xu2020theory} quantifies uncertainty differences across groups within a model family, highlighting the impact of model selection. We next replicate Xu et al.'s \cite{xu2020theory} metric formulations, propose a new metric, and outline their estimation within the $\mathcal{V}$-information framework. The next subsection introduces DispaRisk, a framework for improving fairness analysis in ML pipelines by assessing model class, usable information, and disparate outcomes.

\subsubsection{$\mathcal{V}$-information framework.} Xu et al. \cite{xu2020theory} introduces the $\mathcal{V}$-information framework to estimate \textit{usable information} withim a family of models $\mathcal{V}$. A first formulated concept is the predictive conditional $\mathcal{V}$-entropy, which represents the minimum achievable expected negative log-likelihood to predict $Y$ given $X$ using models from the predictive family $\mathcal{V}$. Formally:

\begin{definition}[Predictive conditional $\mathcal{V}$-\emph{entropy}]
\label{def:cond_v_entropy}
For a family $\mathcal{V}$ of models, the conditional $\mathcal{V}$-entropy\footnote{In this article, conditional $\mathcal{V}$-entropy is referred to as $\mathcal{V}$-entropy.} of $Y$ given $X$ is defined as:
\begin{equation}
    \label{eq:h_x}
    H_{\mathcal{V}}(Y | X) = \inf_{h \in \mathcal{V}} \mathbb{E}_{x,y \sim X,Y}[-\log_2 h[x](y)]
\end{equation}
\end{definition}

The infimum in Equation \ref{eq:h_x} is attained by finding the function $h \in \mathcal{V}$ that minimizes the expected negative log-likelihood,\footnote{With $\log_2$, the measure is in bits; for nats, use $\log_e$.}. Measuring a model class's uncertainty in predicting $Y$ from $X$ requires identifying its best-performing model. Unlike Shannon entropy, $\mathcal{V}$-entropy depends on $\mathcal{V}$, providing distinct uncertainty measures across model classes, making it valuable for comparing predictive capacities -- a key focus of our study.

While $\mathcal{V}$-entropy aggregates uncertainty over the dataset, bias assessment requires analyzing specific data slices, such as demographic differences. To address this, we propose Pointwise $\mathcal{V}$-entropy (\texttt{P}$\mathcal{V}$\texttt{E}) to quantify instance-level uncertainty within a model family $\mathcal{V}$. Formally:

\begin{definition}[Pointwise $\mathcal{V}$-entropy (\texttt{P}$\mathcal{V}$\texttt{E})]
For a family $\mathcal{V}$, and an instance represented by the tuple $(x, y)$, the pointwise $\mathcal{V}$-entropy (\texttt{P}$\mathcal{V}$\texttt{E}) is defined as:
\begin{equation}
\texttt{P}\mathcal{V}\texttt{E}(x\mapsto y) = - \log_2 h[x](y)
\end{equation}
    where $h \in \mathcal{V}$ such that $\mathbb{E}[-\log_2 h[X](Y)] = H_{\mathcal{V}}(Y|X)$.
\end{definition}

Higher \texttt{P}$\mathcal{V}$\texttt{E} values indicate greater uncertainty, meaning models within $\mathcal{V}$ struggle to predict the instance accurately. \texttt{P}$\mathcal{V}$\texttt{E} complements \texttt{PVI} \cite{ethayarajh22difficulty}, which estimates usable information by comparing predictions with and without $x$. In contrast, \texttt{P}$\mathcal{V}$\texttt{E} focuses on the remaining uncertainty when $x$ is given, simplifying estimation and reducing estimation costs. %Analyzing \texttt{P}$\mathcal{V}$\texttt{E} across demographic groups helps identify disparities and instances requiring closer fairness analysis.

\subsubsection{Estimating $\mathcal{V}$-entropy and \texttt{P}$\mathcal{V}$\texttt{E}.}
\label{sec:v_information_estimate}

The $\mathcal{V}$-entropy can be empirically estimated on a finite dataset $\mathcal{D}$ of $n$ instances as:

\begin{align}
\label{eq:emp_inf}
    \hat{H}_\mathcal{V}(Y|X;\mathcal{D}) & = \inf_{h \in \mathcal{V}} \frac{1}{n} \sum_{x_i, y_i\in\mathcal{D}} -\log_2 h[x_i](y_i) \\
                                        & = \inf_{h \in \mathcal{V}} \frac{1}{n} \sum_{x_i, y_i\in\mathcal{D}} \texttt{P}\mathcal{V}\texttt{E}(x_i\mapsto y_i) 
\end{align}
where the infimum $h \in \mathcal{V}$ is approximated using cross-entropy loss to minimize the negative log-likelihood of $Y$ given $X$ \cite{xu2020theory,ethayarajh2022understanding}. The approximation of $\hat{H}_{\mathcal{V}}$ is achieved by training or fine-tuning a pretrained model following Algorithm \ref{algo:v_entropy_estimation}, which extends \cite{ethayarajh2022understanding} to focus on $\mathcal{V}$-entropy and \texttt{P}$\mathcal{V}$\texttt{E}. The algorithm splits data into training and held-out sets, using the latter to estimate $\hat{H}_{\mathcal{V}}$ and \texttt{P}$\mathcal{V}$\texttt{E}. Since estimation is based on finite data, results may deviate from true $\mathcal{V}$-entropy. Xu et al. \cite{xu2020theory} provide Probably Approximately Correct (PAC) bounds, showing that larger datasets and simpler $\mathcal{V}$ yield tighter bounds.

\begin{algorithm}[tbp]
    \caption{$\mathcal{V}$-entropy and \texttt{P}$\mathcal{V}$\texttt{E}}
    \label{algo:v_entropy_estimation}
    \begin{algorithmic}[1]
        \Require $\mathcal{D}_{train}=\{(x_i,y_i)\}_{i=1}^k$, $\mathcal{D}_{held-out}=\{(x_i, y_i)\}_{i=k+1}^n$, and family $\mathcal{V}$
        \Ensure $\hat{H}_\mathcal{V}$ and \texttt{P}$\mathcal{V}$\texttt{E} estimates.
        \State $h \gets$ fine-tune $\mathcal{V}$ on $\mathcal{D}_{train}=\{(x_i,y_i)\}_{i=1}^k$
        \State $\hat{H}_\mathcal{V}(Y|X) \gets 0$
        \For{$(x_i, y_i) \in \mathcal{D}_{held-out}$}
            \State $\hat{H}_\mathcal{V}(Y|X) \gets \hat{H}_\mathcal{V}(Y|X)- \frac{1}{n-k}\log_2 h[x_i](y_i)$
            \State \texttt{P}$\mathcal{V}$\texttt{E}$(x_i\mapsto y_i) \gets -\log_2 h[x_i](y_i)$
        \EndFor
    \end{algorithmic}
\end{algorithm}

\section{DispaRisk}
\label{sec:disparisk}
\subsection{$\mathcal{V}$-information in disparity assessment}
We propose to assess unfairness by comparing $\mathcal{V}$-entropy for predicting $Y$ across advantaged and disadvantaged groups. These uncertainty differences are expected to align with fairness metrics in later ML pipeline stages. To this end, we introduce DispaRisk (\texttt{DR}), a framework for computing these differences and analyzing their relationship with disparities.

First, the difference in the $\mathcal{V}$-entropy across groups is approximated by the average \texttt{P}$\mathcal{V}$\texttt{E} on data slices, where each slice is composed of instances belonging to the advantaged or disadvantaged group. Formally:

\begin{equation}
\label{eq:dif_per_group}
    \texttt{DR}(\mathcal{D}_a, \mathcal{D}_d | \mathcal{V}) = \frac{1}{|\mathcal{D}_a|} \sum_{x,y \in \mathcal{D}_a} \texttt{P}\mathcal{V}\texttt{E}(x\mapsto y) - \frac{1}{|\mathcal{D}_d|}\sum_{x,y \in \mathcal{D}_d} \texttt{P}\mathcal{V}\texttt{E}(x\mapsto y)
\end{equation}
where \texttt{DR} (DispaRisk) approximates uncertainty differences, with $\mathcal{D}_a$ and $\mathcal{D}_d$ representing dataset slices for advantaged and disadvantaged groups, respectively. Since $\mathcal{V}$-entropy is defined over the entire dataset, \texttt{DR} serves as a computationally efficient alternative, requiring only one model per group instead of a full dataset estimation.

Depending on the fairness notion, \texttt{DR} can be computed over specific data slices. For instance, under EQOPP in \textit{separation}, where disparity is assessed only for the positive class, \texttt{DR} is computed over $Y=1$ as follows:

\begin{equation}
\label{eq:dif_per_group_eqopp}
    \begin{split}
        \texttt{DR}(\mathcal{D}_{a,y=1}, \mathcal{D}_{d,y=1} | \mathcal{V}) = \frac{1}{|\mathcal{D}_{a,y=1}|} \sum_{x,y \in \mathcal{D}_{a,y=1}} \texttt{P}\mathcal{V}\texttt{E}(x\mapsto y) \\ - \frac{1}{|\mathcal{D}_{d,y=1}|}\sum_{x,y \in \mathcal{D}_{d,y=1}} \texttt{P}\mathcal{V}\texttt{E}(x\mapsto y)
    \end{split}
\end{equation}
where $\mathcal{D}_{a,y=1}$ and $\mathcal{D}_{d,y=1}$ represent dataset slices for advantaged and disadvantaged groups with target label $y=1$. Thus, when comparing \texttt{DR} with DEMP, we use Equation \eqref{eq:dif_per_group}, and for EQOPP, we use Equation \eqref{eq:dif_per_group_eqopp}.

\subsection{The relationship between \texttt{DR} and fairness notions}
We now examine the relationship between \texttt{DR} and the fairness notions of \textit{separation} and \textit{independence}. Higher uncertainty implies that models in $\mathcal{V}$ are less confident in predicting $Y$ from $X$. When $\mathcal{V}$-entropy is high, models in $\mathcal{V}$ tend to rely on guessing, favoring the majority class in $Y$. Consequently, instances from the group with the highest average \texttt{P}$\mathcal{V}$\texttt{E} are more likely to be predicted as the majority class. How does this affect disparities in $\mathcal{V}$? In the following, we outline rules of thumb to address this question.

\subsubsection{\texttt{DR} and \textit{separation} through EQOPP.}
EQOPP measures the difference in true positive rates between advantaged and disadvantaged groups. By the definition of \texttt{DR} in Equation \eqref{eq:dif_per_group_eqopp}, higher absolute \texttt{DR} values should positively correlate with EQOPP. The reasoning is that greater \texttt{DR} differences indicate that the group with higher average \texttt{P}$\mathcal{V}$\texttt{E} experiences greater uncertainty in predicting $Y$ from $X$, leading to more inaccurate predictions and a lower true positive rate. This results in higher EQOPP values, reflecting greater disparities under \textit{separation}. Based on this analysis, we establish the following rules of thumb, demonstrated in the experiments: \textit{For \textbf{higher} absolute values of \texttt{DR}, \textbf{higher} levels of disparities under EQOPP are expected.}

% \smallskip
% \colorbox{yellow!15}{%
%     \begin{minipage}{.9\textwidth}
%         \begin{center}
%         \textit{For \textbf{higher} absolute values of \texttt{DR}, \\ \textbf{higher} levels of disparities under EQOPP are expected.}
%         \end{center}
%     \end{minipage}%
% }

\subsubsection{\texttt{DR} and \textit{independence} through DEMP.}
DEMP measures the difference in positive ratios between advantaged and disadvantaged groups. To determine whether higher \texttt{DR} values from Equation \eqref{eq:dif_per_group} correspond to higher or lower positive ratios for the group with greater uncertainty—and thus the expected DEMP levels—we identify two key dataset characteristics.

The first is the \textbf{majority class in the target}, which helps predict whether the higher-uncertainty group will receive a higher or lower positive rate. Since \texttt{DR} implies that the group with higher average \texttt{P}$\mathcal{V}$\texttt{E} is more likely to be predicted as the majority class, the relationship between \texttt{DR} and DEMP depends on whether the majority class is positive or negative. If the majority class is positive, the higher-uncertainty group is expected to receive higher positive ratios, leading to greater disparities under DEMP. Thus, a second rules of thumb is: \textit{For \textbf{higher} absolute values of \texttt{DR}, \textbf{higher} levels of disparities under DEMP are expected}.

Conversely, if the \textbf{majority class is negative}, the relationship is reversed. The group with higher uncertainty is now less likely to be predicted as positive, reducing the difference in positive ratios. Therefore, the third rules of thumb is defined as: \textit{For \textbf{higher} absolute values of \texttt{DR}, \textbf{lower} levels of disparities under DEMP are expected}.

\subsection{Benefits of \texttt{DR}}
Our simple yet effective approach offers two key benefits. First, it aligns with fairness notions by accounting for the dependency between labels and sensitive attributes. \texttt{DR} translates fairness concepts into the $\mathcal{V}$-entropy framework, where fairness implies uncertainty differences close to zero, ensuring equal \textit{usable information} across groups and reducing disparities. However, as we will show, this holds only under certain conditions and disparity notions.

Second, $\mathcal{V}$-entropy enables pipeline-dependent metrics for model selection. Since it is defined over $\mathcal{V}$, this set can be tailored to the models used in the ML pipeline, making \texttt{DR} context-specific rather than dataset- or model-specific. A more granular approach could involve multiple $\mathcal{V}$ sets, each representing different model families, allowing for comparable metrics across model types. The following sections demonstrate how DispaRisk enhances disparity risk assessment through a thorough analysis.

\section{Experiments}
\subsection{Machine Learning Pipelines}
We assess disparity risks in three ML pipelines using datasets KDD, FACET, and Hate Speech, denoted as $\mathcal{D}^{kdd}$, $\mathcal{D}^{facet}$, and $\mathcal{D}^{hs}$. Each dataset $\mathcal{D}$ includes input features $X$, sensitive attribute $S$, and target $Y$ for learning $h:X\mapsto Y$. While $S$ is excluded from mapping, it remains available for fairness analysis. We now describe each dataset.

The KDD Census-Income dataset ($\mathcal{D}^{kdd}$) originates from the 1994–1995 U.S. Census Bureau surveys, containing 41 demographic and employment-related variables for 299,285 individuals. It is used to classify whether a person earns more than $50K$ per year, with sensitive attributes such as age, sex, and race.%\footnote{\url{https://archive.ics.uci.edu/ml/datasets/Census-Income\%2B(KDD)}}

The FACET dataset ($\mathcal{D}^{facet}$) is a benchmark from Meta AI for evaluating vision model fairness \cite{gustafson2023facet}. It includes 32,000 images labeled with demographic (e.g., perceived gender presentation) and person-related attributes (e.g., \textit{lawman}, \textit{nurse}), covering 50,000 people. We extract a dataset of 50,000 images (one per person) using provided bounding boxes, along with a binary masculine gender attribute and person-related class labels.%\footnote{\url{https://ai.meta.com/datasets/facet/}}

The Hate Speech dataset ($\mathcal{D}^{hs}$) by Davidson et al. \cite{davidons2017hate} contains 24,802 tweets labeled as \textit{hate speech}, \textit{offensive}, or \textit{neither}. We augment it with demographic dialect predictions from Blodgett et al. \cite{blodgett2016demographic}, estimating dialect proportions for African-American, Hispanic, White, and other groups per tweet.

Following fairness conventions for binary classification, we define disadvantaged group membership using sensitive attributes and the positive class based on target variables. Table \ref{tab:adv_positive} summarizes these criteria. We transform the sensitive attribute, assigning $1$ to disadvantaged groups and $0$ otherwise. Likewise, the target variable is set to $1$ for positive class instances and $0$ for all others.

\begin{table}[t]
    \centering
    \caption{Disadvantage group and positive class from the sensitive attribute and target variable for each ML pipeline.}
    \begin{tabular}{lllll}
         \toprule
         ML Pipeline & Sensitive & Disadvantage & Target & Positive class \\ \midrule
         $\mathcal{D}^{kdd}$ & sex & female & income & $>50K$ \\
         \multirow{2}{*}{$\mathcal{D}^{facet}$} & \multirow{2}{*}{gender} & \multirow{2}{*}{non-masculine} & \multirow{2}{*}{person-related} & lawman \\
                                                & & & & nurse \\ 
        $\mathcal{D}^{hs}$ & dialectal & african-american & harrasement & non-harrasement \\
        \bottomrule
    \end{tabular}
    \label{tab:adv_positive}
\end{table}

\subsection{Disparity Risk Assessments}
We conduct two approaches to evaluate disparity risks in each ML pipeline: (1) a \textit{baseline} using popular \textit{dataset-focused} metrics from literature, and (2) an approach using \texttt{DR} and comparing with popular \textit{model-focused} metrics.

\subsubsection{Baseline.}
\label{sec:baseline}
We use \textit{data-focused} metrics—CIm, DPL, KL, $r_{\phi}$, and Matthews Correlation Coefficient—to assess bias \cite{hardt2021amazon,jurman2012comparison}. Table \ref{tab:pre_metrics} presents the results, highlighting varying bias levels across datasets. For $\mathcal{D}^{kdd}$, we observe slight over-representation of the disadvantaged group, with a moderate negative correlation ($r_{\phi}=-0.159$) between sensitive attributes and labels and a higher positive rate for the male socio-demographic group (DPL$>0$). In $\mathcal{D}^{hs}$, bias is strongest for the positive class (\textit{no\_harassment}), where DPL, $r_{\phi}$, and KL reach the highest absolute values, indicating that the advantaged group has more tweets labeled as \textit{no\_harassment} than disadvantaged one. $\mathcal{D}^{facet}$ shows a strong over-representation of the advantaged group. Based on $r_{\phi}$ and DPL, the \textit{lawman} class has a higher positive rate for the advantaged group, while \textit{nurse} exhibits the opposite trend.

\begin{table}[t]
\caption{\textit{Data-focused} metrics computed from datasets. Higher values indicate stronger relationships between sensitivities and labels.}
    \begin{center}
    \begin{tabular}{lccccc}
    \toprule
    Pipeline & Class & CIm & DPL & $r_{\phi}$ & KL  \\ \midrule
     $\mathcal{D}^{kdd}$ & $>50k$ & -0.04 & 0.08 & -0.16 & 0.07 \\ 
     $\mathcal{D}^{hs}$& \textit{no\_harassment} & -0.06 & 0.25 & -0.34 & 0.33 \\ 
     \multirow{2}{*}{$\mathcal{D}^{facet}$} & \textit{lawman} & \multirow{2}{*}{0.34} & 0.06 & -0.10 & 0.03 \\ 
                                            & \textit{nurse} & & -0.04 & 0.12 & 0.02 \\ \bottomrule
    \end{tabular}
    \label{tab:pre_metrics}
    \end{center}
\end{table}

While these results offer valuable insights, they provide a global perspective, lacking the granularity needed to analyze specific model types within each ML pipeline. The following section applies \textit{DispaRisk} to enable a more nuanced assessment of potential biases in ML pipelines.

\subsubsection{DispaRisk in Practice.}
\label{sec:infobias_in_practice}
DispaRisk is applied in three steps: (1) constructing model families based on the intended models for the ML pipeline, (2) estimating \texttt{DR}, and (3) analyzing results to generate insights. %While our pipelines are hypothetical, we assume model selections to construct relevant families. In practice, model families would be defined by real-world constraints, implementation needs, and model preferences.

\paragraph{(1) Construction of families $\mathcal{V}$.} For each hypothetical ML pipeline, we define model families based on assumed preferences. For $\mathcal{D}^{kdd}$, we construct five Feedforward Neural Network (FNN) families with different activation functions: no activation (linear), ReLU \cite{zeiler2013relu}, LeakyReLU \cite{he2015delving}, Sigmoid, and GELU \cite{hendrycks2016gelu}. For $\mathcal{D}^{hs}$, we analyze transformer-based families: BERT \cite{devlin2019bert}, RoBERTa \cite{liu2019roberta}, GPT2 \cite{radford2019language}, BART \cite{lewis2020bart}, and DeBERTa \cite{he2021deberta}. For $\mathcal{D}^{facet}$, we employ popular vision model families: VGG \cite{simonyan2015vgg}, Inception \cite{szegedy2016inception}, DenseNet \cite{huang2017densenet}, MobileNet \cite{howard2019mobile}, and VisionTransformer \cite{dosovitskiy2021vit}. Model families are identified using activation functions or model architecture names as subscripts. For example, $\mathcal{V}{leaky_relu}$ represents FNNs with LeakyReLU, and $\mathcal{V}{gpt2}$ denotes the GPT2 family.

\paragraph{(2) \texttt{DR} Estimates.} To estimate uncertainty differences via \texttt{DR} (Section \ref{sec:disparisk}), we follow this protocol for each ML pipeline using dataset $\mathcal{D}^{(p)}$ and family $\mathcal{V}_i$:

\begin{enumerate}
    \item[(1)] Split $\mathcal{D}^{(p)}$ into $\mathcal{D}_{train}^{(p)}$ and $\mathcal{D}_{held-out}^{(p)}$ sets at 80/20 ratio.
    \item[(2)] Approximate the infimum $h \in \mathcal{V}_i$ by training or fine-tuning a pretrained model using cross-entropy loss (Step 1 of Algorithm \ref{algo:v_entropy_estimation}).
    \item[(3)] Estimate $H_{\mathcal{V}_i}$ and \texttt{P}$\mathcal{V}$\texttt{E} following Steps 2–6 of Algorithm \ref{algo:v_entropy_estimation}.
\end{enumerate}

Since the most computationally powerful model in $\mathcal{V}$ often attains the infimum (Definition \ref{def:cond_v_entropy}), this weakens the PAC bound \cite{xu2020theory}, requiring overfitting prevention. To mitigate this, we create a validation set $\mathcal{D}^{(p)}_{val}$ by sampling 10\% of $\mathcal{D}^{(p)}_{train}$ and evaluate estimates per epoch. Models are trained/fine-tuned for 5 epochs with a learning rate of $5e-5$ and batch size 32. If overfitting arises, we lower the learning rate to $5e-6$, halve the batch size, and rerun Algorithm \ref{algo:v_entropy_estimation}. We use the AdamW optimizer \cite{loshchilov2017decoupled} with a linear scheduler\footnote{Minimum learning rate set to 0.} for all experiments. %Appendix \ref{sec:appendix_dr_estimates_training} describes in detail the results from training for \texttt{DR} estimates.

\begin{figure}[t]
    \centerline
    {\includegraphics[width=.8\linewidth]{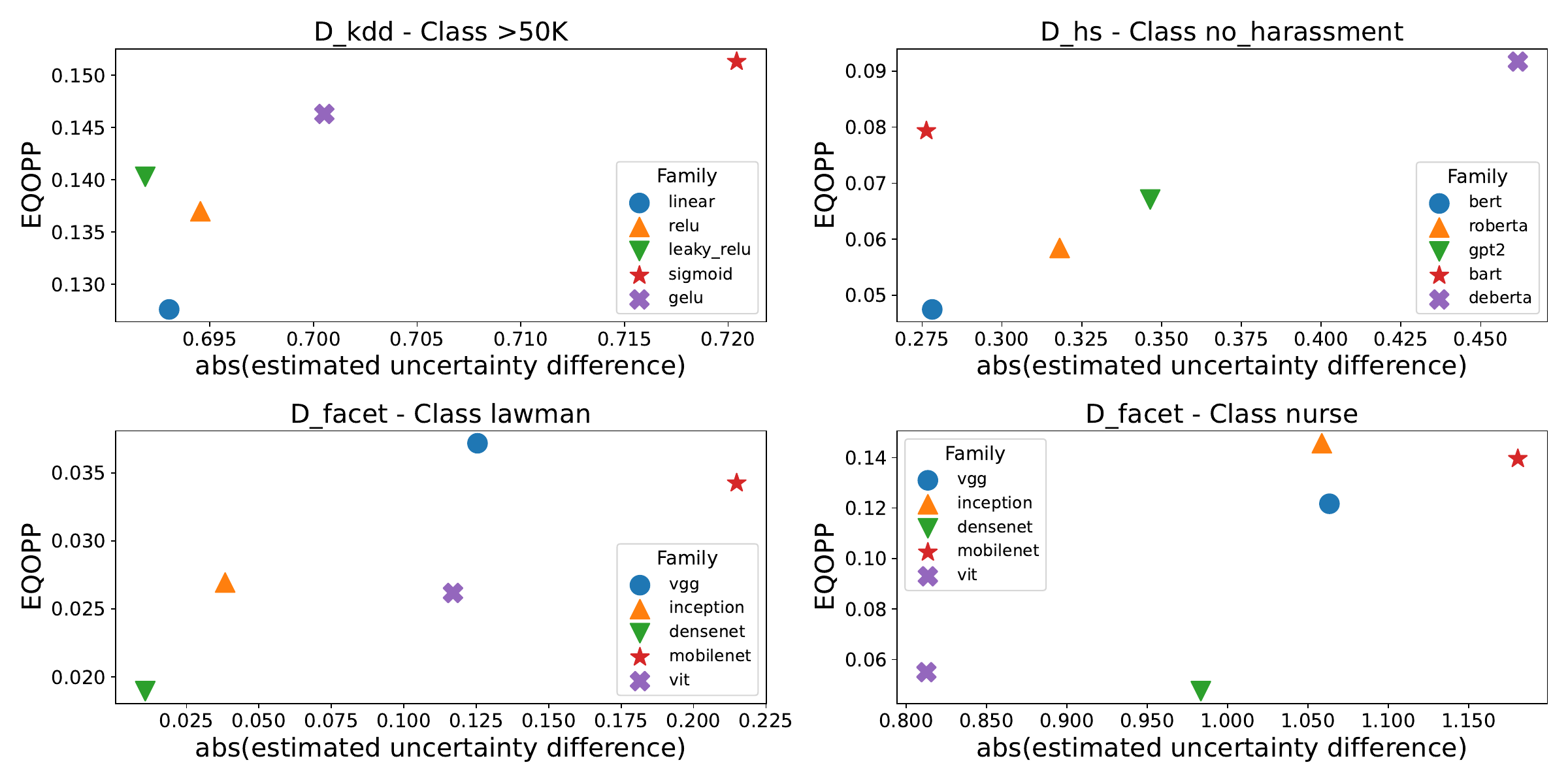}}
    \caption{Average EQOPP disparity in family versus estimated uncertainty difference through $\texttt{DR}(\mathcal{D}_{a,y=1},\mathcal{D}_{d,y=1})$ for each family $\mathcal{V}_i$.}
    \label{fig:eqopp_vs_uncertainty_diff}
\end{figure}

\begin{figure}[t]
    \centerline
    {\includegraphics[width=.8\linewidth]{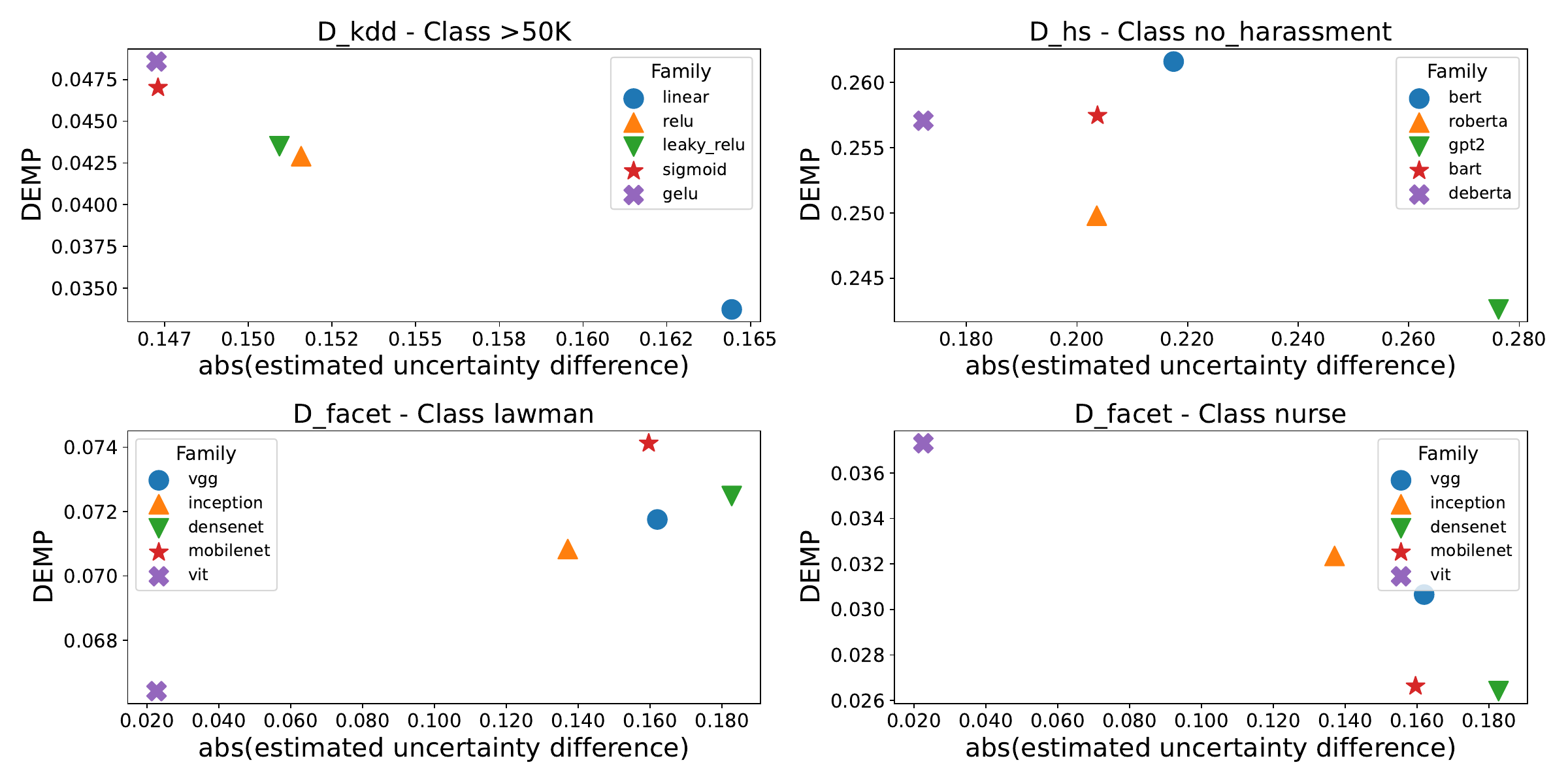}}
    \caption{Average DEMP in predictive families versus estimated uncertainty difference through $\texttt{DR}(\mathcal{D}_{a},\mathcal{D}_{d})$ for each family $\mathcal{V}_i$.}
    \label{fig:demp_vs_uncertainty_diff}
\end{figure}

\paragraph{(3) Assessing Fairness Through Usable Information.} We use the estimates to address two key questions that \textit{data-focused} metrics alone cannot answer:

\textbf{(Q.1) Which model families in the ML pipeline are more likely to replicate or exacerbate biases?} To investigate this, we simulate later ML pipeline stages and compare estimated \texttt{DR} with observed disparities, identifying model families prone to higher bias reproduction. For example, in the $\mathcal{D}^{kdd}$ pipeline, we first estimate \texttt{DR} for each family. Next, we train FNNs with varying hidden layers from each family and compute average disparity levels using EQOPP and DEMP. In parallel, we estimate \texttt{DR} for each family. Finally, we compare uncertainty difference estimates with observed disparities to evaluate whether \texttt{DR} effectively signals model families more prone to exacerbating biases, as inferred in Section \ref{sec:disparisk}. Applying this protocol across all ML pipelines, we obtain the results shown in Figures \ref{fig:eqopp_vs_uncertainty_diff} and \ref{fig:demp_vs_uncertainty_diff}, which depicts the relationship between \texttt{DR} estimates and EQOPP/DEMP across different pipelines and model families. % In the following paragraph we perform analysis revealing distinct patterns across datasets that helps to answer question (\textbf{Q.1}).

Figure \ref{fig:eqopp_vs_uncertainty_diff} shows the relationship between absolute \texttt{DR}($\mathcal{D}_{a,y=1},\mathcal{D}_{d,y=1}$) values and average EQOPP. The observed trend confirms the rule of thumb from Section \ref{sec:disparisk} across all ML pipelines%: higher absolute \texttt{DR} values correlate with greater disparities in EQOPP
, validating \texttt{DR} as a predictor of disparity risks for future models in downstream tasks. Given this, we derive the first insight for \textbf{(Q.1)}: \textit{the model families most prone to higher disparities under \textit{separation} notions are $\mathcal{V}_{sigmoid}$, $\mathcal{V}_{deberta}$, and $\mathcal{V}_{mobilenet}$ for the $\mathcal{D}^{kdd}$, $\mathcal{D}^{hs}$, and $\mathcal{D}^{facet}$ pipelines, respectively.}

Figure \ref{fig:demp_vs_uncertainty_diff} illustrates the relationship between \texttt{DR}($\mathcal{D}_a,\mathcal{D}_d$) and DEMP. To analyze these results and address \textbf{Q.1}, we follow a structured approach: (1) identify the majority class, (2) evaluate whether the observed trends align with the expected \texttt{DR}-DEMP relationship from Section \ref{sec:disparisk}, and (3) synthesize insights to answer \textbf{Q.1}. Applying this approach, we find that for $\mathcal{D}^{kdd}$ the negative class (income $<50k$) is the majority, suggesting an inverse relationship between absolute \texttt{DR} and DEMP (Section \ref{sec:disparisk}). Figure \ref{fig:demp_vs_uncertainty_diff} confirms this, with higher absolute \texttt{DR} values corresponding to lower DEMP. Models in $\mathcal{V}_{gelu}$ and $\mathcal{V}_{sigmoid}$ show higher disparity risks. Additionally, in $\mathcal{D}^{hs}$ the majority class is \textit{harassment} (negative class), indicating a similar inverse \texttt{DR}-DEMP relationship as in $\mathcal{D}^{kdd}$. Consistent with Section \ref{sec:disparisk}, models in $\mathcal{V}_{deberta}$ are more prone to disparities under \textit{independence}. Finally, for $\mathcal{D}^{facet}$ the majority class is \textit{lawman}. From Figure \ref{fig:demp_vs_uncertainty_diff}, the rule of thumb (Section \ref{sec:disparisk}) that an inverse \texttt{DR}-DEMP relationship is expected for the \textit{nurse} class and a direct one for \textit{lawman} is confirmed. Models in $\mathcal{V}_{densenet}$ are more prone to disparities for \textit{lawman}, while $\mathcal{V}{vit}$ shows similar tendencies for \textit{nurse}.

\textbf{(Q.2) Why might these model types produce disparate outcomes?} The rules of thumb not only help identify high-risk model families but also explain why these families contribute to disparities in later pipeline stages. For example, in $\mathcal{V}_{sigmoid}$, EQOPP is higher because the group with a higher average \texttt{P}$\mathcal{V}$\texttt{E} is less likely to be correctly predicted, leading to a lower true positive rate. Similarly, DEMP is higher as the lower average \texttt{P}$\mathcal{V}$\texttt{E} differences shows that models are reflecting dataset biases seen in Table \ref{tab:pre_metrics}. This pattern generalizes across model families and ML pipelines.

To further explore these disparities, we analyze which features contribute to the computed average \texttt{P}$\mathcal{V}$\texttt{E} in each model family. We select the riskiest families per ML pipeline and identify key features by measuring uncertainty reduction when a feature is added to the input space. Specifically, we compare \texttt{P}$\mathcal{V}$\texttt{E} when feature $i$ is masked using transformation $\tau_i$ versus when $x$ is complete. The transformations applied in each pipeline are: setting feature $i$ to $0$ in $\mathcal{D}^{kdd}$, replacing word $i$ with a blank space in $\mathcal{D}^{hs}$, and setting specific pixel sets to $0$ in $\mathcal{D}^{facet}$. Formally, this uncertainty reduction is measured as follows:

\begin{align}
    \texttt{UR}(\mathcal{D} | \mathcal{V}, \tau_i) & =  \frac{1}{|\mathcal{D}|}\sum_{x,y\in \mathcal{D}} \texttt{P}\mathcal{V}\texttt{E}(\tau_i(x)\mapsto y)-\texttt{P}\mathcal{V}\texttt{E}(x\mapsto y) \\
    & = \frac{1}{|\mathcal{D}|} \sum_{x,y\in \mathcal{D}} -\log_2 h[x_{\neg i}](y) + \log_2 h[x](y)
\end{align}
where \texttt{UR} represents Uncertainty Reduction, $\tau_i(x)$ denotes the transformation process that masks feature $i$, and $x_{\neg i}$ is the resulting output. We use $h \in \mathcal{V}$ such that $\mathbb{E}[-\log_2 h[X](Y)] = H_{\mathcal{V}}(Y|X)$. Higher \texttt{UR} values for feature $i$ indicate its importance for models in $\mathcal{V}$ to accurately predict the target variable. Notably, we use the same infimum of $\mathcal{V}$ for \texttt{P}$\mathcal{V}$\texttt{E} with both $\tau_i(x)$ and the unmasked input to avoid the computational overhead of determining a separate infimum for each masked feature. While this simplification has limitations, which will be discussed in Section \ref{sec:discussion}, the primary goal here is to demonstrate how \textit{DispaRisk} extends beyond identifying risky models to offer deeper insights into potential disparities. In the following paragraphs we apply this approach to all ML pipelines.

\begin{figure}[t]
    \centering
    \includegraphics[width=.7\linewidth]{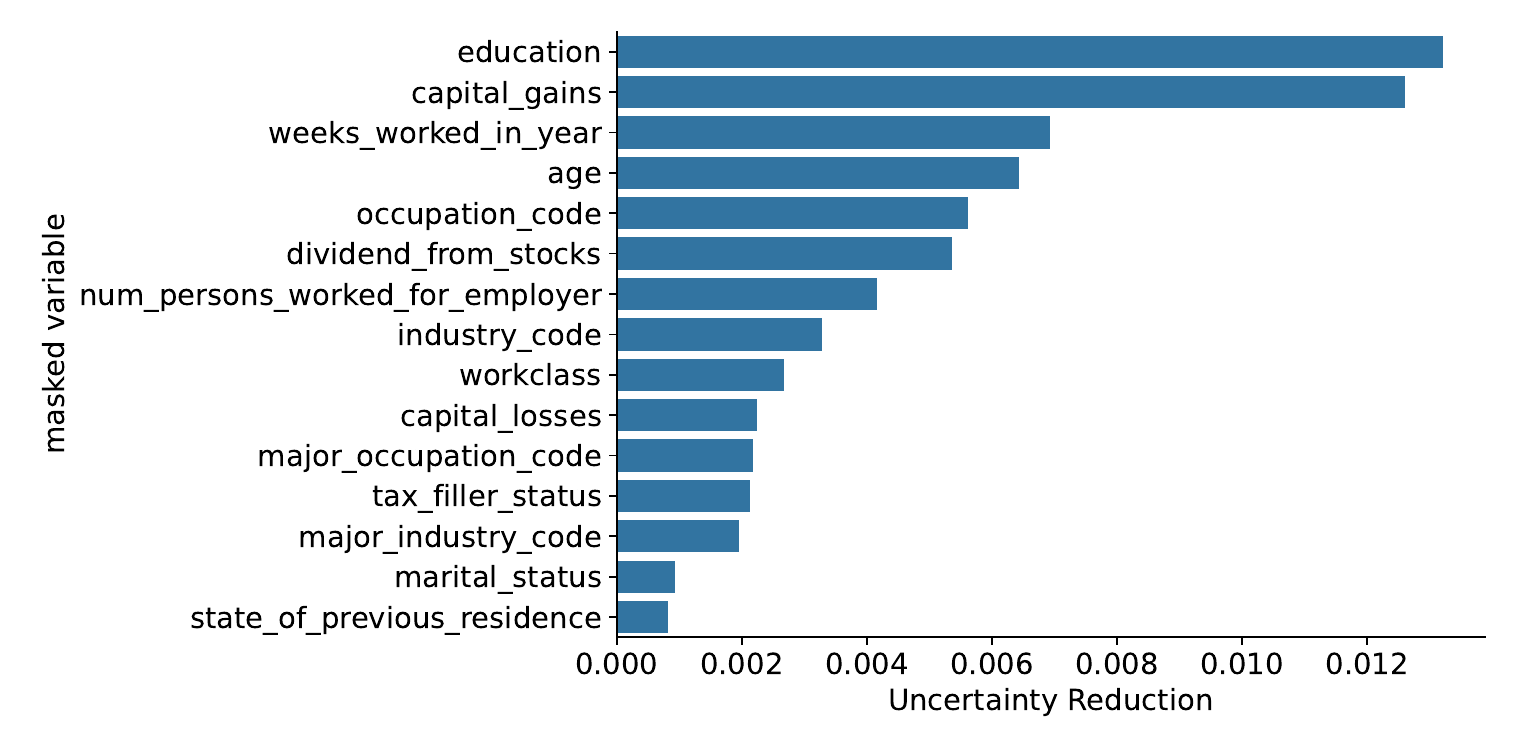}
    \caption{Top $15$ \texttt{UR} of features over advantage group (male) for the $\mathcal{V}_{sigmoid}$ family in the $\mathcal{D}^{kdd}$ ML pipeline.}
    \label{fig:feature_kdd}
\end{figure}

In $\mathcal{D}^{kdd}$, the positive DPL indicates higher labeled positive rates for the advantaged group, a disparity replicated in $\mathcal{V}_{sigmoid}$ due to its lower \texttt{DR}. Thus, we compute \texttt{UR} on $\mathcal{D}_a$. Figure \ref{fig:feature_kdd} shows the top 15 most relevant features by \texttt{UR} for $\mathcal{V}_{sigmoid}$, the highest-risk family. The main contributors to disparity risks are \texttt{education}, \texttt{capital\_gains}, \texttt{weeks\_worked\_in\_year}, \texttt{age}, and \texttt{occupation}, with \texttt{education} and \texttt{capital\_gains} providing the greatest uncertainty reduction. Thus, complementing \textbf{(Q.1)}, uncertainty differences are largely attributed to these variables in $\mathcal{V}_{sigmoid}$. This might suggest insights such as careful preprocessing of these features to mitigate bias or further considerations on these variables during model constructions.

\begin{figure}[t]
    \centering
    \includegraphics[width=.7\linewidth]{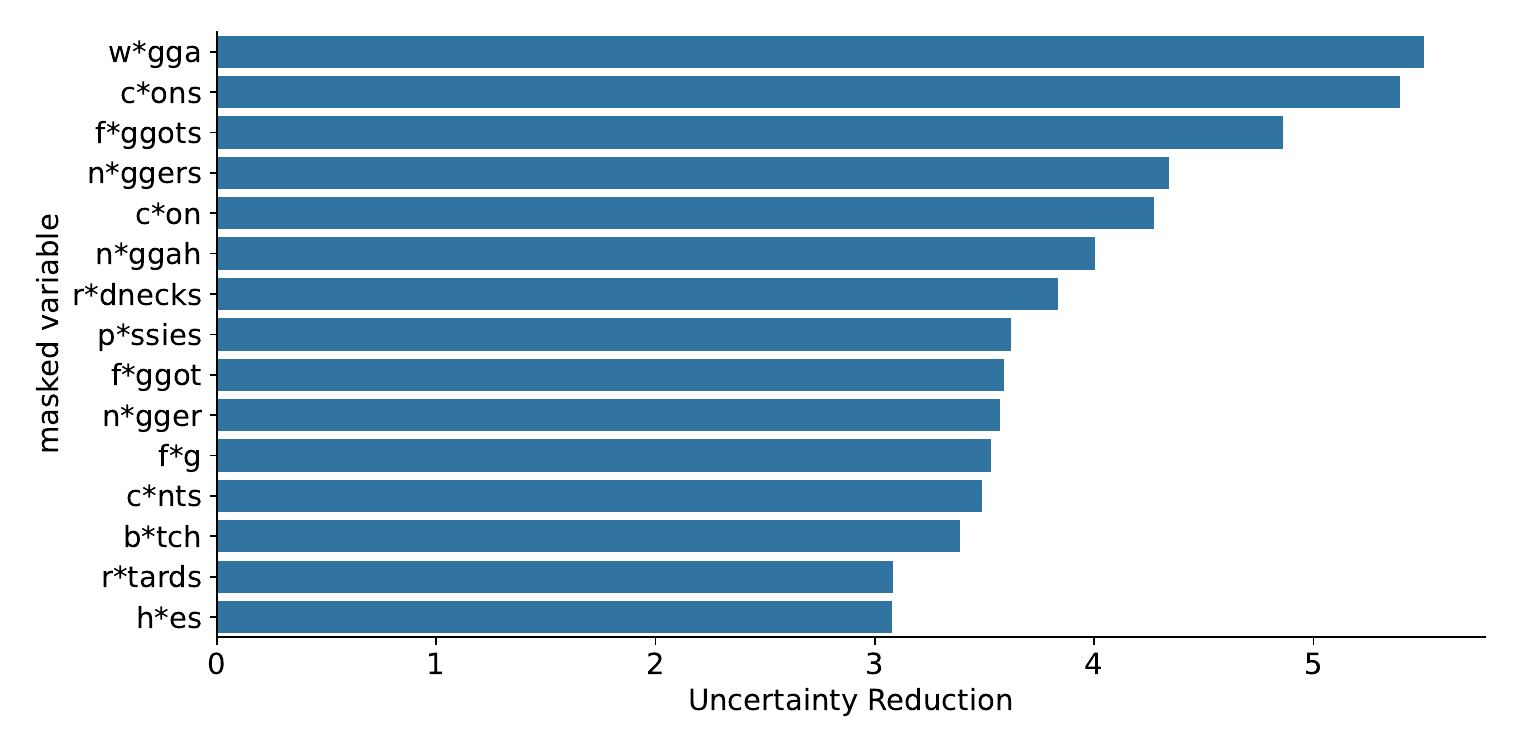}
    \caption{Top $15$ \texttt{UR} of words over advantage group (Not-African-American dialect) for the $\mathcal{V}_{deberta}$ family in the $\mathcal{D}^{hs}$ ML pipeline. Words are modified to avoid exposition of inappropriate text.}
    \label{fig:feature_hs}
\end{figure}

For the $\mathcal{D}^{hs}$ pipeline, we analyze two word sets: (1) the most relevant words for each target class identified by Ethayarajh et al. \cite{ethayarajh2022understanding} and (2) a manually curated list of problematic words for $\mathcal{D}^{hs}$. For each word $i$, we compute \texttt{UR} using a subsample of texts containing $i$, capturing its impact on uncertainty reduction when present versus absent. Given the \texttt{DR} values and following the approach for $\mathcal{D}^{kdd}$, we compute \texttt{UR} on the advantaged group due to the positive DPL. Figure \ref{fig:feature_hs} presents the top 15 words in $\mathcal{V}_{deberta}$, identified as the highest-risk family in \textbf{(Q.1)}. The analysis shows that racial and homophobic slurs most significantly reduce uncertainty in $\mathcal{V}_{deberta}$ for the advantaged group. This indicates that a specific set of biased terms largely drives uncertainty differences and, consequently, disparity risks in this predictive family.

\begin{figure}[t]
    \centering
    \includegraphics[width=.7\linewidth]{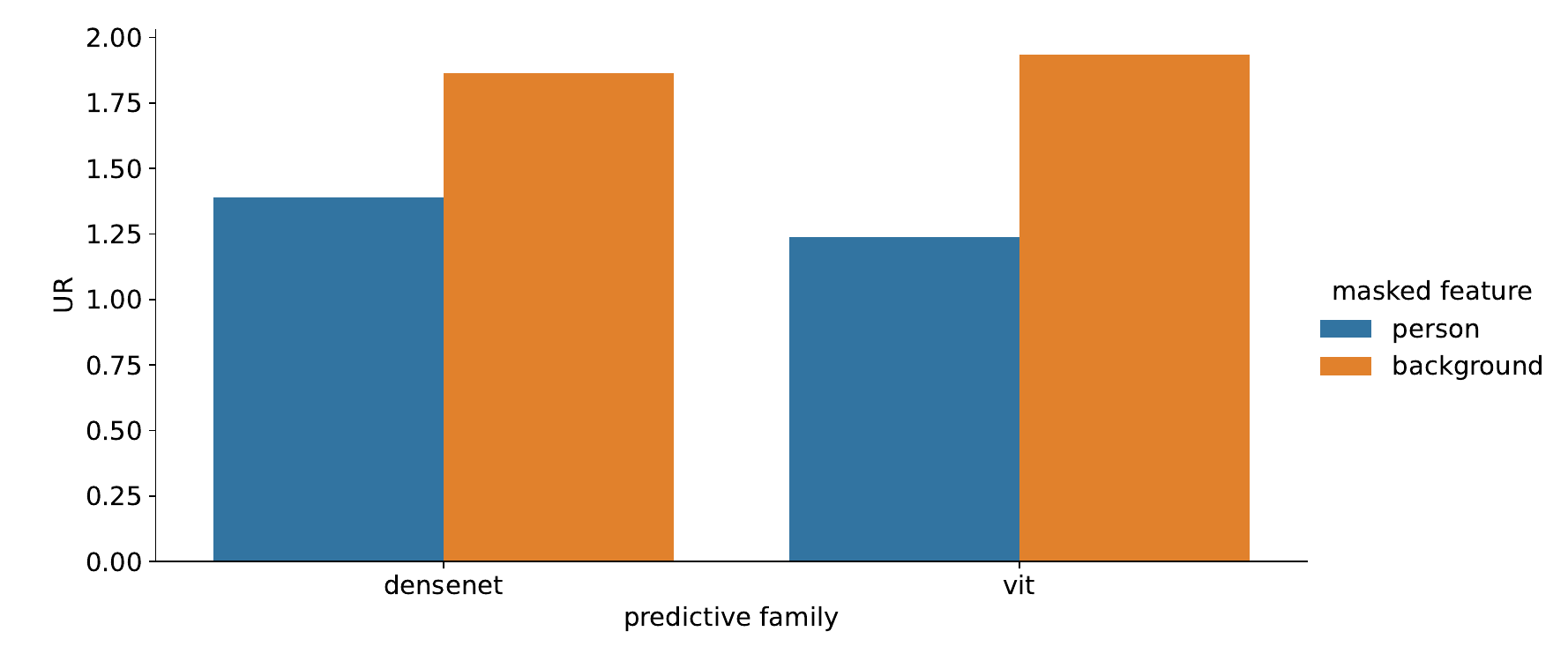}
    \caption{\texttt{UR} of feature \texttt{person} and \texttt{background} over disadvantage group (Female) for the $\mathcal{V}_{densenet}$ and $\mathcal{V}_{vit}$ families in the $\mathcal{D}^{facet}$ ML pipeline.}
    \label{fig:feature_facet}
\end{figure}

Finally, for the $\mathcal{D}^{facet}$ dataset, we focus on the disadvantaged group, which had the highest average \texttt{P}$\mathcal{V}$\texttt{E} in the estimated \texttt{DR}. We analyze $\mathcal{V}_{densenet}$ and $\mathcal{V}_{vit}$, identified as the highest-risk families for the \texttt{lawman} and \texttt{nurse} classes, respectively. We examine the image background and the person to determine which contributes more to uncertainty reduction, explaining the disparity risks identified earlier. Figure \ref{fig:feature_facet} shows that in both families, the background has the highest uncertainty reduction. This reinforces that image backgrounds significantly impact the elevated uncertainty of the disadvantaged group, driving disparity risks. These findings help narrow the focus on key features when analyzing biases.

\section{Discussion}
\label{sec:discussion}
\textit{DispaRisk} bridges the gap between \textit{data-focused} and \textit{model-focused} bias detection methods. While it operates on datasets like \textit{data-focused} techniques, it also provides insights that are both \textit{data-centric} and \textit{model-family-aware}, making it distinct from traditional approaches.

A key difference from \textit{model-focused} metrics is its broader applicability. Model-focused metrics evaluate bias based on a specific model's output, limiting generalizability. In contrast, \textit{DispaRisk} estimates metrics over an entire model family $\mathcal{V}$ (Definition \ref{def:cond_v_entropy}), enabling a more generalizable assessment earlier in the ML pipeline. Another advantage of \textit{DispaRisk} is its scalability to high-dimensional data. Unlike mutual information, which struggles with high-dimensional variables, empirical $\mathcal{V}$-metrics remain tractable as dimensionality increases \cite{xu2020theory}. This is especially useful in modern ML, where high-dimensional data is common.

However, \textit{DispaRisk} has limitations. Estimating $\mathcal{V}$-\textit{entropy} requires training or fine-tuning at least one model per family, making it computationally expensive compared to traditional \textit{data-focused} methods. This poses challenges in external audits where resources may be limited. Another limitation is the reliance on properly defining $\mathcal{V}$. A poorly chosen or overly restrictive $\mathcal{V}$ may lead to misleading results or fail to capture relevant biases, introducing subjectivity. Different choices of $\mathcal{V}$ can produce varying outcomes, requiring domain expertise to ensure an appropriate definition. Additionally, $\mathcal{V}$-metric accuracy depends on dataset size and training settings. Smaller datasets can yield less reliable estimates, weakening bias assessment validity. A further challenge lies in \texttt{UR} computation, where selecting feature $i$ is predefined, making it inefficient for high-dimensional settings, such as large vocabularies. This inefficiency makes \textit{DispaRisk} dependent on expert knowledge to identify key features and challenging when assessing ML pipelines with limited data, highlighting areas for future research.

\section{Conclusions}
\textit{DispaRisk} enhances bias assessment in ML pipelines by addressing veracity and value challenges in large datasets. Using the $\mathcal{V}$-entropy framework, it reveals how different model families may amplify societal biases, bridging \textit{data-focused} and \textit{model-focused} metrics while considering computational constraints. Illustrative experiments across diverse datasets demonstrate \textit{DispaRisk}’s effectiveness in identifying disparity sources and explaining bias propagation. By pinpointing areas where biases are amplified, it helps improve dataset quality and fairness in ML applications. Its context-specific assessments make it a valuable tool for regulatory compliance and internal use, ensuring fairness in ML-based systems.

Future research on \textit{DispaRisk} can explore several key directions to enhance its applicability and impact. One promising avenue is refining its estimation methods to improve efficiency, enabling faster assessments in large-scale ML pipelines. Another important direction is adapting \textit{DispaRisk} for evolving ML architectures, ensuring its relevance as models become more complex and diverse. Finally, investigating its role in AI governance and regulatory compliance can help establish standardized fairness auditing practices, fostering greater transparency and accountability in machine learning systems.

% \begin{credits}
% \subsubsection{\ackname} A bold run-in heading in small font size at the end of the paper is
% used for general acknowledgments, for example: This study was funded

% \subsubsection{\discintname}
% It is now necessary to declare any competing interests or to specifically
% state that the authors have no competing interests. Please place the
% statement with a bold run-in heading in small font size beneath the
% (optional) acknowledgments,
% for example: The authors have no competing interests to declare that are
% relevant to the content of this article. Or: Author A has received research
% grants from Company W. Author B has received a speaker honorarium from
% Company X and owns stock in Company Y. Author C is a member of committee Z.
% \end{credits}

\bibliographystyle{splncs04}
\bibliography{references}
%% Note that this preceding line implies that you store your BibTeX references in a file called 'mybibliography.bib'. If you instead store your references in a file with a different name, for instance 'references.bib', the preceding line should read '\bibliography{references}'. Whatever you do, DO NOT put the file name extension .bib inside the \bibliography command; this will trip up LaTeX compilers. 
%
% If you do not want to use BibTeX, you can also type up the bibliography exactly as you see fit, using the following structure:
% Note that this number 8 reserves an amount of space (equal to the natural width of the given number) for the label of your references; if you have more than 9 references, you will want to change this number to 18. If you have more than 19 references, this number is best changed to 88. If you have more than 99 references, I salute you.




% LNCS Homepage, \url{http://www.springer.com/lncs}, last accessed 2023/10/25
% \end{thebibliography}
\end{document}